\documentclass[twoside]{article}
\usepackage{amsthm}
\newcommand{\bm}[1]{\mathbf{#1}}
\newcommand{\w}[1]{\mathcal{#1}}
 
\newcommand{\bs}[1]{\boldsymbol{#1}}
\newtheorem{theorem}{Proposition}
\usepackage[round]{natbib}

\newtheorem{definition}{Definition}[section]
\usepackage{amsfonts}
\newcommand\numberthis{\addtocounter{equation}{1}\tag{\theequation}}
\usepackage{enumitem}
\setlist{nosep} 
\newtheorem{assumption}{Assumption}
\usepackage{graphicx}
\newcommand{\mycomment}[1]{}
\usepackage{placeins}
\usepackage{float}
\usepackage{hyperref}
\usepackage{xr-hyper} 

\usepackage{color}

\makeatletter
\newcommand*{\addFileDependency}[1]{
\typeout{(#1)}
\@addtofilelist{#1}
\IfFileExists{#1}{}{\typeout{No file #1.}}
}\makeatother

\newcommand{\myexternaldocument}[1]{%
\externaldocument{#1}%
\addFileDependency{#1.tex}%
\addFileDependency{#1.aux}%
}
\myexternaldocument{supplement}

\usepackage[accepted]{aistats2024}

\begin{document}

%

%

\twocolumn[

\aistatstitle{Trigonometric Quadrature Fourier Features for Scalable Gaussian Process Regression}

\aistatsauthor{ Kevin Li \And Max Balakirsky \And  Simon Mak}
\aistatsaddress{ Duke University \And   \And Duke University} ]

\begin{abstract}
Fourier feature approximations have been successfully applied in the literature for scalable Gaussian Process (GP) regression. In particular, Quadrature Fourier Features (QFF) derived from Gaussian quadrature rules have gained popularity in recent years due to their improved approximation accuracy and better calibrated uncertainty estimates compared to Random Fourier Feature (RFF) methods. However, a key limitation of QFF is that its performance can suffer from well-known pathologies related to highly oscillatory quadrature, resulting in mediocre approximation with limited features. We address this critical issue via a new Trigonometric Quadrature Fourier Feature (TQFF) method, which uses a novel non-Gaussian quadrature rule specifically tailored for the desired Fourier transform. We derive an exact quadrature rule for TQFF, along with kernel approximation error bounds for the resulting feature map. We then demonstrate the improved performance of our method over RFF and Gaussian QFF in a suite of numerical experiments and applications, and show the TQFF enjoys accurate GP approximations over a broad range of length-scales using fewer features.
\end{abstract}

\section{INTRODUCTION}
Gaussian Processes (GPs) \citep{gpml} are a popular class of Bayesian non-parametric models. Unfortunately, for large sample sizes $n \gg 1000$, the $\mathcal{O}(n^3)$ cost for GP training and prediction can be prohibitive in applications. There has been much work on addressing this critical issue, including inducing points \citep{titsias_spgr, hensman_svgp, fitc}, nearest-neighbor approximations \citep{wu_vnn, jian_cao_vnn, katzfuss_nn}, iterative numerical methods \citep{gpytorch, gp_sgd, wang_big_gp}, and divide-and-conquer approaches \citep{poe_deisenroth, mixture_moe}.


Our paper will focus the use of Fourier feature approximations \citep{rahimi}, which have shown promise in recent work. The key idea is to construct a low-rank approximation of the covariance matrix for a stationary GP, using a finite set of Fourier features dervied from the kernel's spectral density. Fourier approximations have three key advantages for GP regression: they allow for nice approximation error bounds, perform inference using a finite feature basis, and exploit the spectral representation of covariance kernels. As such, such approximations are increasingly popular in broad applications, including generalized Bayesian quadrature \citep{rff_quad}, deep GPs \citep{cutajar_dgp}, latent variable models \citep{rfflvm}, differential privacy \citep{dubey_proviate, dp_rff}, federated learning \citep{federated_rff}, Bayesian optimization, \citep{qff_weighted, Mutny}, and spatial statistics \citep{ton}. 

For GPs, there has been two main directions for Fourier feature approximation. The first direction, Random Fourier Features (RFF; \citealp{rahimi}), uses Monte Carlo sampling to generate features. The integration of RFF for GP regression is easy-to-implement and scales nicely to high dimensions. However, RFF methods are known to suffer from a phenomenon called \textit{variance starvation}, which can lead to poorly calibrated uncertainty estimates and erratic predictive mean behavior \citep{wilson_posterior, wilson_pathwise, Mutny, wang_starve}. 

The second direction, Quadrature Fourier Features (QFF; \citealp{dao_quad, Mutny, gl_qff}), aims to alleviate variance starvation via a deterministic Gaussian quadrature rule for the Fourier transform integral. This has been successfully applied for lower-dimensional GP applications, including Bayesian optimization  \citep{dp_rff, dubey_proviate, Mutny, bo_Qff_heavytailed}, robust inference \citep{qff_robustness}, spatial-temporal data \citep{gl_qff}, and derivative modeling \citep{sleip}. Compared to RFF, the deterministic quadrature rules in QFF permit quicker error decay and can thus avoid variance starvation. In practice, however, achieving this improved performance over RFF can require an undesirably large number of features, particularly with small length-scales for the underlying GP \citep{Mutny, gl_qff}. 

This paper proposes a new Trigonometric Quadrature Fourier Features (TQFF) that addresses the aforementioned limitations of existing RFF and QFF methods. We show that the use of Gaussian quadrature rules in QFF (which rely on polynomial interpolants) can lead to poor performance with small length-scales when approximating the highly oscillatory Fourier transform. Motivated by this, the TQFF uses a novel quadrature rule that relies on a \textit{trigonometric} interpolant, tailored specifically for the desired Fourier transform. In doing so, we show empirically that the TQFF enjoys accurate GP approximations over a broad range of length-scales using fewer features. \mycomment{Sections \ref{sec:back} and \ref{sec:related} outline existing and related work. Section \ref{sec:trig} presents the proposed quadrature rule, and investigates its associated approximation error bounds. Section \ref{sec:exp} then empirically demonstrates the improved performance of TQFF over Gaussian QFF and RFF, in a suite of numerical experiments.}


\mycomment{
\textbf{Contributions} 1) we analyze of the shortcomings of current Gaussian QFF methods for GP regression 2) propose a new method for obtaining Fourier features by employing a novel trignometrically exact quadrature rule 3) provide approximation error bounds for the new feature map 4) show empirically that TQFF outperforms RFF and QFF methods for low-dimensional GP regression tasks. 
}

\section{BACKGROUND}
\label{sec:back}

\subsection{Gaussian Process Regression}

A Gaussian process $f(\cdot)$ is a stochastic process for which its evaluation on any finite subset of inputs follows a multivariate Gaussian distribution. Here, we assume the standard regression set-up, with observed data $\w{D} = \{\bm{x}_i, y_i \}_{i=1}^n$ where $\bm{x}_i \in \mathbb{R}^d$ and $y_i \in \mathbb{R}$. The standard GP regression framework then follows:
\begin{align*}
&y_i = f(\bm{x}_i) + \epsilon_i, \quad \epsilon_i \overset{i.i.d.}{\sim} \w{N}(0, \sigma^2),
\end{align*}
where $f(\cdot) \sim \w{GP}(0, k_{\bs{\Theta}}(\cdot, \cdot))$ follows a zero-mean GP with kernel $k_{\bs{\Theta}}$. Here, $\bs{\Theta}$ consists of all kernel hyperparameters, including length-scale and scale parameters. Training of these kernel hyperparameters $\bs{\Theta}$ and the noise variance $\sigma^2$ can proceed via maximizing the following log-marginal likelihood of $\bm{y} = (y_i)_{i=1}^n$:
\begin{align*}
\log p(\bm{y}| \bm{X}, \bs{\Theta}) = \log \phi(\bm{y}; \mathbf{0}, \bm{K}_{\mathbf{X}\mathbf{X}} + \mathbf{I}_{n \times n}\sigma^2),
\end{align*}
where $\phi(\cdot; \boldsymbol{\mu},\Sigma)$ is the multivariate Gaussian density. Here, $\bm{K}_{\mathbf{X}\mathbf{X}} \in \mathbb{R}^{n\times n}$ is the covariance matrix with $(i,k)$-th entry $k_{\bs{\Theta}}(\bm{x}_i, \bm{x}_k)$. Model inference and prediction thus requires the inversion of a $n \times n$ matrix, which requires $\w{O}(n^3)$ operations and can thus be prohibitive for large $n \gg 1000$.

\mycomment{This paper will focus on the sqaured exponential kernel covariance which has form 
\begin{align*}
k_{\bs{\theta}}(\bm{x}_{0}, \bm{x}_{1}) = \nu \exp\left(-\frac{1}{2}\sum_{j=1}^d \frac{ (\bm{x}^{(j)}_0 - \bm{x}^{(j)}_1)^2}{\bs{\theta}^{(j)}}\right)
\end{align*}
Where $\bm{x}^{(j)}$ represents the $j$th entry of vector $\bm{x}$.  Here $\bs{\Theta} = \{\bs{\theta}, \nu \}$.}

\mycomment{
Conditional on the observed data $\w{D}$, the predictive distribution of new observation $y_{*}$ and location $\bm{x}_{*}$ is gaussian with expectation and variance: 
\begin{align*}
&\mbox{E}(y_{*}| \bm{x}_{*}, \w{D}) = \bm{k}_{*}^T(\bm{K}_{\bm{X}\bm{X}} + I_{n \times n}\sigma^2)^{-1} \bm{y} \\
&\mbox{V}(y_{*}|\bm{x}_{*}, \w{D}) = k_{\bs{\theta}}(\bm{x}_{*}, \bm{x}_{*}) - \bm{k}_{*}^T(\bm{K}_{\bm{X}\bm{X}} + I_{n \times n}\sigma^2)^{-1}\bm{k}_{*} + \sigma^2
\end{align*}
Where $\bm{k}_{*} \in \mathbb{R}^{n}$ such that $\bm{k}_{*}[i] = k_{\bs{\theta}}(\bm{x}_{*}, \bm{x}_{i})$. Training and prediction require inversion of the $n \times n$ matrix requires a prohibitive $\w{O}(n^3)$ operations.  
}

\subsection{Gaussian Quadrature}
We provide a brief review of classical Gaussian quadrature; for details, see \cite{num_text}. Gaussian quadrature approximates integrals of the form:
\[\int_{a}^b p(\omega) h(\omega) d\omega, \quad -\infty \leq a < b \leq \infty,\]
where $p(\omega)$ is the \textit{weight} function and $h(\omega)$ is the \textit{integrand}. Different weight functions give rise to different quadrature rules. Gaussian quadrature makes the approximation $h(\omega) \approx P_{L-1}(\omega)$ where $P_{L-1}(\omega)$ is an $L-1$ degree polynomial interpolating $h(\omega)$ at $L$ quadrature nodes $\{\omega_l\}_{l=1}^L$, $\omega_l \in [a, b]$. With this approximation, the quadrature rule becomes:
\begin{align}
&\int_{a}^b p(\omega) h(\omega) d\omega \approx \sum_{l =1}^L a_l h(\omega_l) := Q_L(h), \label{eqn:quad}
\end{align}
where $a_l = \int_{a}^b h(\omega) t_l(\omega) d\omega$ and $t_l(\omega)$ are the $L-1$-degree Lagrange interpolating polynomials \citep{num_text}. This quadrature rule thus requires the set of quadrature nodes $\{\omega_l\}_{l=1}^L$ and quadrature weights $\{a_l\}_{l=1}^L$. Gaussian quadrature rules select these nodes such that Equation \eqref{eqn:quad} is exact for polynomial $h(\omega)$ of degree up to $2L-1$. However, the accuracy of such quadrature rules (and polynomially-exact quadrature rules in general) depends on how well the polynomial interpolant approximates the integrand $h(\omega)$. 

\mycomment{$p(\omega) = \exp(-\omega^2)$ gives Gauss Hermite quadrature, while $p(\omega) = 1$ gives to Gauss-Legendre. The accuracy of Gaussian and polynomially exact quadrature in general depend entirely on how well the integrand can be approximated by polynomials.} 

The above 1-dimensional quadrature rules can directly be extended for multiple dimensions via tensor products; details of this in Appendix \ref{sup:tensor_product}. We do note that, with tensor product rules, the number of nodes grows exponentially with dimensions, which can limit such approaches to problems in low dimensions or with low-dimensional structure.

\subsection{Fourier Feature Approximation}
We now briefly review the general Fourier feature approximation approach. Using Bochner's theorem \citep{rudin2017fourier}, any stationary covariance function $k_{\bs{\Theta}}(\bm{x}, \bm{x}') =k_{\bs{\Theta}}(\bm{x} -\bm{x}') $ can be represented as the Fourier transform of a non-negative measure $p_{\bs{\Theta}}(\omega)$:
\begin{align*}
 k_{\bs{\Theta}}(\bm{x} - \bm{x}') &= \int p_{\bs{\Theta}}(\bs{\omega}) \exp(i\bs{\omega}^T(\bm{x} - \bm{x}')) d\bs{\omega} \numberthis \label{eqn:rff_integral}\\
 & =  \int p_{\bs{\Theta}}(\bs{\omega}) \cos(\bs{\omega}^T(\bm{x} - \bm{x}')) d\bs{\omega},
\end{align*}
where the last line assumes both data and kernel are real-valued. Fourier feature methods then use the following finite feature approximation:
\begin{align*}
\small
&k_{\bs{\Theta}}(\bm{x}, \bm{x}') \approx \sum_{s=1}^S a_s \cos(\bs{\omega}_s^T (\bm{x} - \bm{x}')) = \bs{\Phi}(\bm{x})^T\bs{\Phi}(\bm{x}'), \numberthis \label{eqn:rff_approx} 
\normalsize
\end{align*}
where $\bs{\Phi}(\bm{x}) \in \mathbb{R}^{2S}$ and 
\begin{align*}
\bs{\Phi}(\bm{x})^{(s)} = \begin{cases}
    \sqrt{a_{s}} \cos(\bs{\omega}_s) & \mbox{if} \  1 \leq s \leq S, \\
    \sqrt{a_{s}} \sin(\bs{\omega}_s) & \mbox{if} \  S < s \leq 2S.
\end{cases}
\end{align*}
Letting $\bs{\Lambda} = (\bs{\Phi}(\bm{x}_1), \dots \bs{\Phi}(\bm{x}_n))$, we have $\bm{K}_{\bm{X}\bm{X}} \approx \bs{\Lambda}^T \bs{\Lambda}$. This allows the use of the efficient matrix determinant and inversion updates (e.g., the Woodbury lemmas \citep{hager1989updating}) for GP training and prediction using $\w{O}(S^3 + n)$ runtime and $\w{O}(Sn)$ memory. The number of features $S$ is pre-set based on computational concerns, with 100-1000 features typical in applications \citep{randomized_truncation, sparse_spectrum, Mutny}.

Existing methods for Fourier feature approximation differ in their choice of $\omega_s$ and $a_s$. We review two popular approaches used for GP regression below:
\begin{itemize}[leftmargin=*]
    \item \textbf{Random Fourier Features} (RFF; \citealp{rahimi}) approximate the integral in \eqref{eqn:rff_integral} via Monte Carlo, where $\bs{\omega_s}$ is sampled from $p_{\bs{\Theta}}(\bs{\omega})$ and $a_s = 1/S$ so the estimator in $ \eqref{eqn:rff_approx}$ is a sample average.
    \item \textbf{Gaussian Quadrature Fourier Features} (Gaussian QFF) approximate the integral in \eqref{eqn:rff_approx} via Gaussian quadrature, where $\omega_s$ and $a_s$ are selected from numerical quadrature rules. \cite{Mutny} makes use of Gauss Hermite Fourier feature (GHFF) maps, defining  $p_{\bs{\Theta}}(\omega)$ after a change of variable as the weight function and $h(\omega) = \cos(\bs{\omega}^T(\bm{x} - \bm{x}'))$ as the integrand. Such an approach, however, is restricted to GPs with the squared exponential (SE) kernel. \cite{gl_qff} make use of Gauss Legendre Fourier feature (GLFF) maps, where after sufficient truncation of the integral, the weight function is constant and $h(\omega) = p_{\bs{\Theta}}(\bs{\omega}) \cos(\bs{\omega}^T(\bm{x} - \bm{x}'))$ is the integrand. Such choices have significant impact on approximation accuracy, as we shall see next.

\end{itemize}

\begin{figure}[t]
  \centering
  \includegraphics[width=\columnwidth]{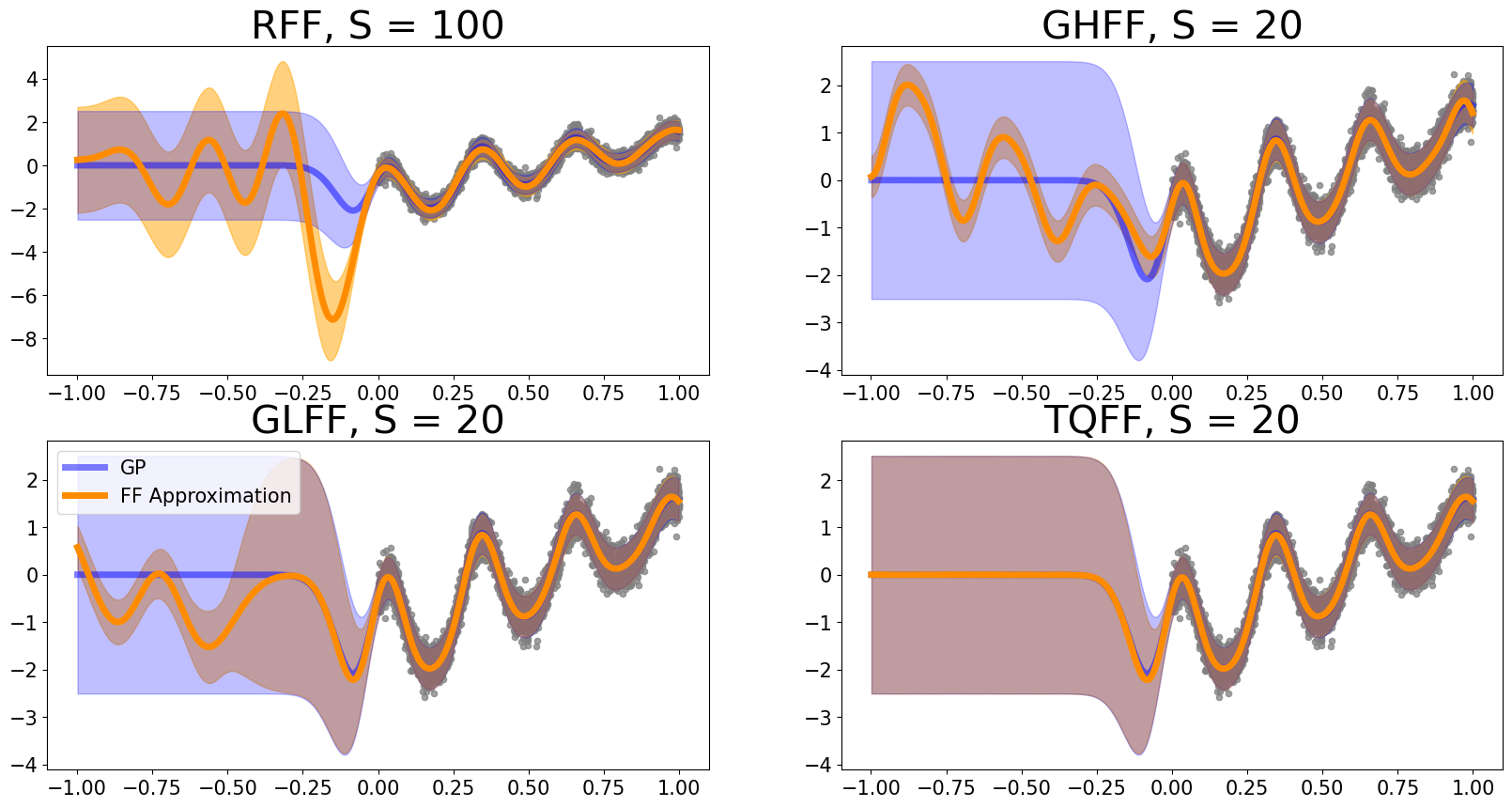}
  \caption{Predictions and 95\% predictive intervals from various Fourier feature approximation methods with $S$ features (orange), compared to a fitted full GP (blue). Models share the same hyperparameters optimized from the full GP. }
  \label{fig:toy_example}
\end{figure}

In what follows, we will exploit the symmetry of Gaussian quadrature rules and the Fourier integrand to eliminate half the nodes when constructing GLFF and GHFF maps. As such, we will derive Gaussian QFF maps using $S$ features from the corresponding $2S$-point quadrature rule. 

\subsection{Potential Drawbacks of Gaussian QFF}
\label{sec:adv_disadv}

\begin{figure}
    \centering
    \includegraphics[width=\columnwidth, height = 4.5cm]{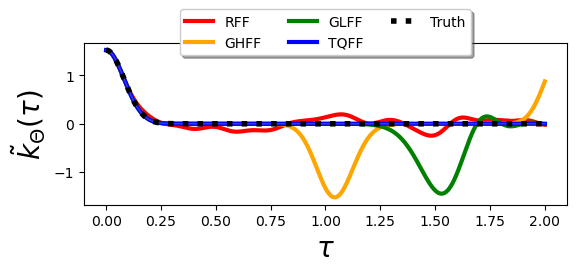}
    \caption{Kernel approximations using various Fourier feature approximation methods, as a function $\tau = x - x'$ for the toy example in Figure \ref{fig:toy_example}.}
    \label{fig:kernel_approx}
\end{figure}

We illustrate the advantages and drawbacks of Gaussian QFF via a toy example. Figure \ref{fig:toy_example} shows the fits (using $n = 5000$ training samples) from a full GP with the SE kernel and its various Fourier feature approximations\footnote{Experimental details can be found in Appendix \ref{sup:toy}.}. All methods adopt the same hyperparameters $\bs{\Theta}$ obtained via maximization of the full GP marginal likelihood. We see clearly that the RFF and GHFF suffer from the aforementioned ``variance starvation'' in the data sparse region, whereas GLFF performs slightly better. Figure \ref{fig:kernel_approx} shows the corresponding approximations of $k_{\bs{\Theta}}(\tau)$ from each method. RFF has trouble estimating covariances for small $\tau$ due to the slow decay of Monte Carlo error, which results in a severe underestimation of posterior uncertainty. On the other hand, GHFF and GLFF (the deterministic feature maps) return significant errors when estimating covariances for larger $\tau$. Such discrepancies help explain the poor predictions in Figure \ref{fig:toy_example}.

\begin{figure}
    \centering
    \includegraphics[width=\columnwidth]{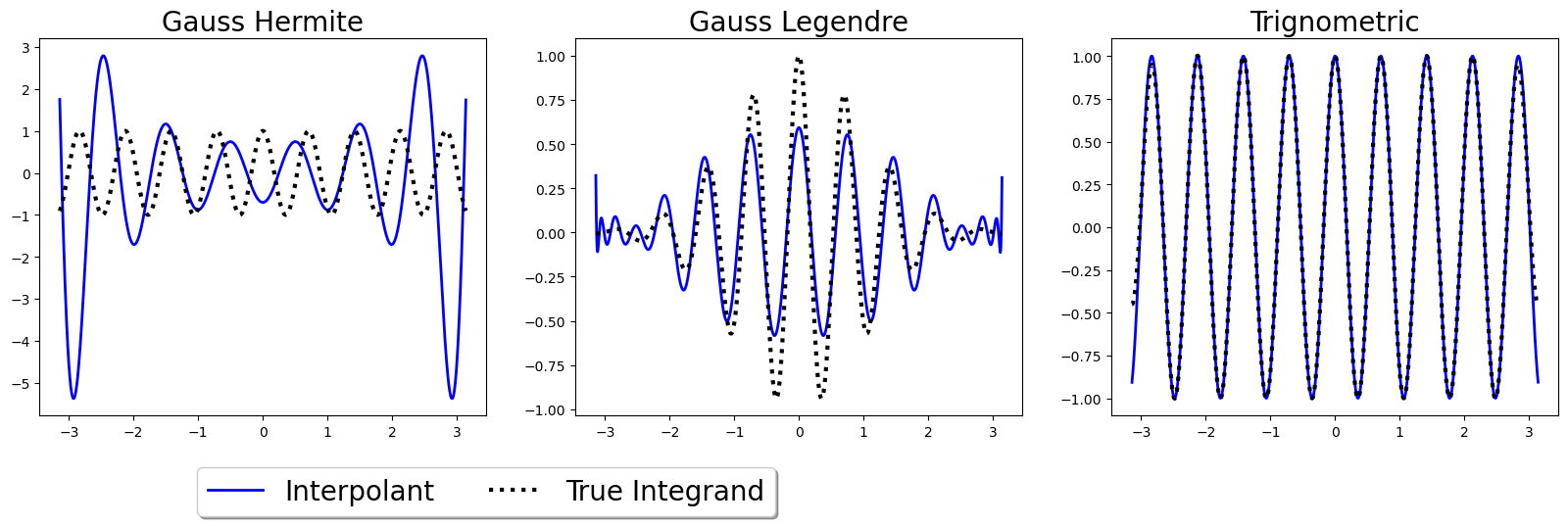}
    \caption{Integrands and interpolants implied by the quadrature rules approximating $k_{\bs{\Theta}}(1.75)$ with $S= 15$ features.}
    \label{fig:interpolant}
\end{figure}

One explanation for the large errors of Gaussian QFF is its reliance on polynomial interpolation to approximate the oscillatory integrand in Equation \ref{eqn:rff_integral}, which becomes increasingly oscillatory with large $\tau$ or low length-scales $\theta$. Figure \ref{fig:interpolant} shows the interpolants and integrands implied by these quadrature feature maps, when approximating $k_{\bs{\Theta}}(1.75)$ with $S=15$ features. The GHFF interpolant clearly yields a poor approximation of the integrand, whereas the GLFF interpolant performs better; this is not surprising, as Gauss-Legendre quadrature incorporates the rapidly decaying spectral density into the integrand, which damps the oscillatory behavior. These observations suggest that QFF can be greatly improved if one factors in the specific quadrature approximation problem of interest. In doing so, we show next that we can retain the Gaussian QFF's ability to express near-zero covariances (i.e., at small $\tau$) without sacrificing large errors at large $\tau$ or at small length-scales.

\mycomment{
\paragraph{QFF Efficiently Approximate Near-Zero Covariances}

To motivate the development of Trigonometric Quadrature Features (TQF), we examine the desirable and undesirable characteristics of current Gaussian QFF methods. We begin by constructing a synthetic function and generating $n = 5000$ training points in the interval $x \in [0, 1]$\footnote{Furhter details of this exercise can be found in the supplentary material}. We fit a full GP with a Squared Exponential kernel to this training set and our goal will be to approximate the GP using Fourier feature methods. Figure \ref{fig:toy_example} shows the resulting approximations along with the full GP. We see that RFF and Gauss-Hermite QFF with $S = 300$ and $S=30$ features respectively both suffer heavily from variance starvation. Interestingly, however Gauss Legendre QFF and our proposed Trigonometric QFF perform significantly better, with the trigneomtric quadrature approximation nearly indistinguishable from the full GP. 

Figure \ref{fig:kernel_approx} shows the values of the approximated kernel covariance $\tilde{k}_{\bs{\Theta}}(\tau)$ for the methods and various $\tau= x - x', x, x' \in [-1, 1]$.  We see that the variance starvation of RFF is caused by \textit{covariance saturation}. The slow decay of monte-carlo error means that RFF feature maps cannot efficiently approximate covariances that are near-zero and therefore overestimate the covariance between the training and test sets. Unfortunately, accurate representation of zero covariance is neccesary to allow the approximated GP to default to the prior, a crucial characteristic of GP uncertainty quantification. 

The Gaussian QFF methods are capable of expressing near-zero covariances with few features, but suffer from periodic large deviations from the truth value as $\tau$ increases.  This leaves us to the key drawback of Gaussian QFF methods for GPs.

\subsection{Gaussian Quadrature is Unsuited for Trignometric Integrals}
The accuracy of Gaussian Quadrature fundamentally depends how effectively the interpolating polynomial approximates the integrand. It is widely known in applied mathematics that this approximation performs poorly when applied to highly oscillatory integrands. We demonstrate this poor performance in Figure \ref{fig:interpolant} where we plot for each quadrature rule the relevant integrand  and  interpolating function used to approximate $k_{\bs{\Theta}}(1.75)$ with  $S = 15$\footnote{We reduce to 15 features, as explicitly calculating Lagrange interpolation is very numerically unstable for $S \geq 20$} Fourier features. We keep the length-scale of the kernel from the previous example. Further details on this example can be found in the appendix. 

We see that the interpolant performs the worst for Gauss Hermite quadrature, which explains the method's poor performance in Figure \ref{fig:toy_example}. This is unsurprising as $cos(k\omega)$ is very oscillatory as $k$ increases and becomes increasingly difficult to approximate with polynomials. Gauss Legendre performs second best, as its incorporates the rapidly decaying spectral density function $p_{\bs{\Theta}}(\omega)$ into the integrand, which results in a tamer interpolation problem. Still, as we see from the results of the trignometric rule, we can do much better if we carefully choose our interpolant. }

\section{RELATED WORK}
\label{sec:related}

\paragraph{Quadrature of highly oscillatory integrals} has been studied extensively in the applied mathematics literature \citep{deano2017computing}. Gaussian quadrature is widely known to be sub-optimal for oscillatory quadrature problems, which has motivated alternate approaches including Filon \citep{filon}, Levin \citep{levin}, and steepest descent quadrature \citep{deano2017computing}. While such methods are highly accurate, these quadrature rules cannot be adapted into viable feature maps for kernel approximation. \cite{milovanovic2006trigonometric, milovanovic2008trigonometric, da2012trigonometric} study quadrature rules that are exact for trigonometric polynomials. However, such rules involve solving complex systems of equations that may not have solutions for many weight functions. Furthermore, guaranteeing exactness for all trigonometric polynomials can be wasteful in our setting, and results in inefficient feature approximation maps. 

\paragraph{Fourier Feature Approximations for GPs} Our work is most related to the GP developments in \cite{Mutny, gl_qff}.  \cite{Mutny} proposed Gauss Hermite QFF for learning GPs in Bayesian optimization. Recently, \cite{gl_qff} made use of Gauss Legendre QFF (along with a rigorous method for choosing sample size dependent hyperparameters) to guarantee spectral equivalence between the full kernel covariance and its low-rank approximation. \cite{randomized_truncation} further analyzed RFF methods, and found they systematically overfit to data. Significant work has been done on RFF methods for non-GP-related kernel approximation tasks. \cite{avron_qmc} analyze error bounds for Quasi Monte Carlo sampled feature maps, while \cite{yu2016orthogonal, munk} propose sampling from restricted geometries to achieve Monte Carlo variance.

\section{TRIGONOMETRIC QUADRATURE FOURIER FEATURES}
\label{sec:trig}

To address the aforementioned limitations of existing RFF and QFF methods, we propose a new Trigonometric Quadrature Fourier Feature (TQFF) approach. The core idea is to derive a quadrature rule via a cosine interpolant specifically tailored for Fourier transform integrand in Equation \eqref{eqn:rff_integral}. We first derive this quadrature rule, then provide its kernel approximation error bounds. We defer all proofs to Appendix \ref{sup:proofs}.

\subsection{Kernel Assumptions}

We first make the following assumptions on the stationary kernel covariance function $k_{\bs{\Theta}}(\cdot, \cdot)$:
\begin{assumption} The stationary kernel covariance function $k_{\bs{\Theta}}(\cdot, \cdot)$ satisfies the following:
\label{assumptions}
\begin{enumerate}[leftmargin=*,label=(\alph*)]
    \item The kernel can be written exactly as:
        \[\small k_{\bs{\Theta}}(\mathbf{x} - \mathbf{x}') = g(\bs{\Theta}) \int p(\bs{\omega}) \exp(i\bs{\omega}^T\bm{D}(\bs{\Theta})(\bm{x} - \bm{x}')) d\bs{\omega} \normalsize\]
    for some scalar function $g(\bs{\Theta})$, matrix-valued function $\bm{D}(\bs{\Theta})$, and density $p(\bs{\omega})$ with no dependency on $\bs{\Theta}$.
    \item The density $p(\bs{\omega})$ in (a) factors over dimensions such that $p(\bs{\omega}) = \prod_{j=1}^d p^{(j)}(\bs{\omega}^{(j)})$.
\end{enumerate}
\end{assumption}
The first assumption states that the kernel is the Fourier dual of a spectral density, and we can perform a change-of-variables such that the density does not depend on kernel hyper-parameters. The second assumption is quite common, and can be found in seminal works \citep{dao_quad, Mutny, avron_qmc}. Both assumptions are satisfied by common kernel choices, including the SE, $1$-d Mat\'ern, and the product-Mat\'ern kernels.

\subsection{Trignometrically Exact Quadrature}

For exposition, we begin with the one-dimensional case, which we will later generalize to higher dimensions. From Assumption \ref{assumptions}, we can write $k_{\Theta}$ as: 
\begin{align*}
&k_{\Theta}(x, x') =g(\Theta)  \int_{-\infty}^{\infty} p(\omega) \exp\left(i\omega 
\left[\frac{x-x'}{\theta} \right] \right) d\omega \\
& \approx g(\Theta)  \int_{-\pi}^{ \pi} p_{\gamma}(\gamma \omega) \cos\left( \omega \gamma \left[\frac{x - x'}{\theta} \right]\right) d\omega, \numberthis \label{eqn:integral}
\end{align*}
where $\theta$ is its length-scale parameter and $\gamma > 0$ is a pre-set truncation parameter. While Gaussian quadrature aims to achieve exact approximation for polynomial integrands, our trigonometrically exact quadrature rules will instead be exact for integrals of the form of Equation \eqref{eqn:integral} when $\gamma \left[ (x - x')/\theta \right]$ is an integer. This leads us to the following definition of a trignometrically exact rule for our use-case:

\begin{definition}[Trigonometric Degree of Exactness]
\label{def:exactness}
 A quadrature rule $Q_S^c(f)$ has trigonometric exactness of degree $K$ with respect to weight function $p_{\gamma}(\gamma \bs{\omega})$ if:
 \begin{align*}
 Q_S^c(\cos(\bs{\omega}^T \bm{k})) =  \int_{ \left[-\pi, \pi \right]^d}  p_{\gamma}(\gamma \bs{\omega}) \cos(\bs{\omega}^T \bm{k} )) d\bs{\omega}
\end{align*}
for $\bm{k} \in \mathbb{N}^d$  and $||\bm{k}||_{\infty} \leq K$.
\end{definition}

We next derive a trignometrically exact rule in one-dimension, by interpolating the integrand $\cos\left( \omega \gamma \left[(x - x')/{\theta} \right]\right)$ using cosine polynomials and integrating the interpolant against the weight function. A cosine polynomial of degree $L$, $p^c_L(\omega)$, has the form $p^c_L(\omega) = b_0 + \sum_{l=1}^L b_l \cos(\omega)^l$. We call a cosine polynomial $p^c_L(x)$ \textit{monic} if $b_L = 1$. The unique cosine polynomial $P_{L-1}^c(\omega)$ of degree $L-1$ that interpolates $f(\omega)$ at $L$ distinct nodes $\{ \omega_l \}_{l=1}^L , \omega_l \in \left[0 , \pi \right)$ has the form  $P_{L-1}^c(\omega) = \sum_{l=1}^L f(\omega_l) t_l^c(\omega)$, where:
\begin{align}
t_l^c(\omega) = \prod_{l=z,l \neq l}^L \frac{\cos(\omega) - \cos(\omega_z)}{\cos(\omega_l) - \cos(\omega_z)} \label{eqn:trig_comp}
\end{align}
Due to the existence of Chebyshev polynomials and uniqueness of the interpolating polynomial, if $f(\omega) = \cos(k \omega )$ and $k \leq L -1$, $k \in \mathbb{N}$, then $P_L^c(\omega) = \cos(k\omega)$. As shown in Figure \ref{fig:interpolant}, this family of cosine polynomials interpolate the Fourier transform integrand much better than polynomial interpolants of similar degrees.

Using this family of interpolants. we can then derive an $L$-point quadrature rule that achieves a trigonometric exactness of degree $2L -1$. We formalize this rule in the following proposition. 
\begin{theorem}[1-d Trigonometric Quadrature]
\label{prop:tqff}
Adopt the conditions in Assumption \ref{assumptions}. Further let $\{q_l^c(\omega)\}_{l=0}^{L}$ be a sequence of orthogonal monic cosine polynomials with degree $l$ such that: 
\begin{align*}
\int_{- \pi}^{ \pi} q_l^c(\omega) q_{l'}^c(\omega) p_{\gamma}(\gamma \omega) d\omega = 0  \quad  \text{if and only if} \  l' \neq l.
\end{align*}
Let $\{\omega_i\}_{i=1}^L$ be the $L$ unique, real-valued zeroes of $q^c_L(\omega)$ in $[0, \pi)$, and define $t_i^c(\omega)$ as in Equation \ref{eqn:trig_comp}. Then, with $a_l = \int_{-\pi}^{\pi} t^c_l(\gamma\omega) p(\gamma\omega) d\omega \geq 0$, the quadrature rule $Q_L^c(f) = \sum_{l=1}^L a_l f(\omega_l)$ has trigonometric exactness of degree $2L-1$.
\end{theorem}

The kernel approximation derived from the $2L - 1$-degree exact trigonometric quadrature rule will be equal to the truncated integral when $\gamma\left[\frac{x - x'}{\theta}\right] \leq 2L - 1$ and is an integer. We note that our choice to only consider exactness for cosine polynomials increases the efficiency of our quadrature. The rules proposed by \cite{milovanovic2006trigonometric, milovanovic2008trigonometric} that are exact for general trigonometric polynomials require $2L$ nodes to achieve degree exactness $2L-1$. 

Unlike rules that guarantee general trigonometric degree of exactness, the nodes and weights that satisfy the conditions of Proposition \ref{prop:tqff} can be computed via classical tools from numerical quadrature. In what follows, we use the the popular Golub-Welsh algorithm \cite{golub1969calculation} to find such quadrature nodes and weights \citep{num_text}. Details on implementation details and computational complexity are provided in supplementary materials. 

\begin{figure}
    \centering
    \includegraphics[width=\columnwidth]{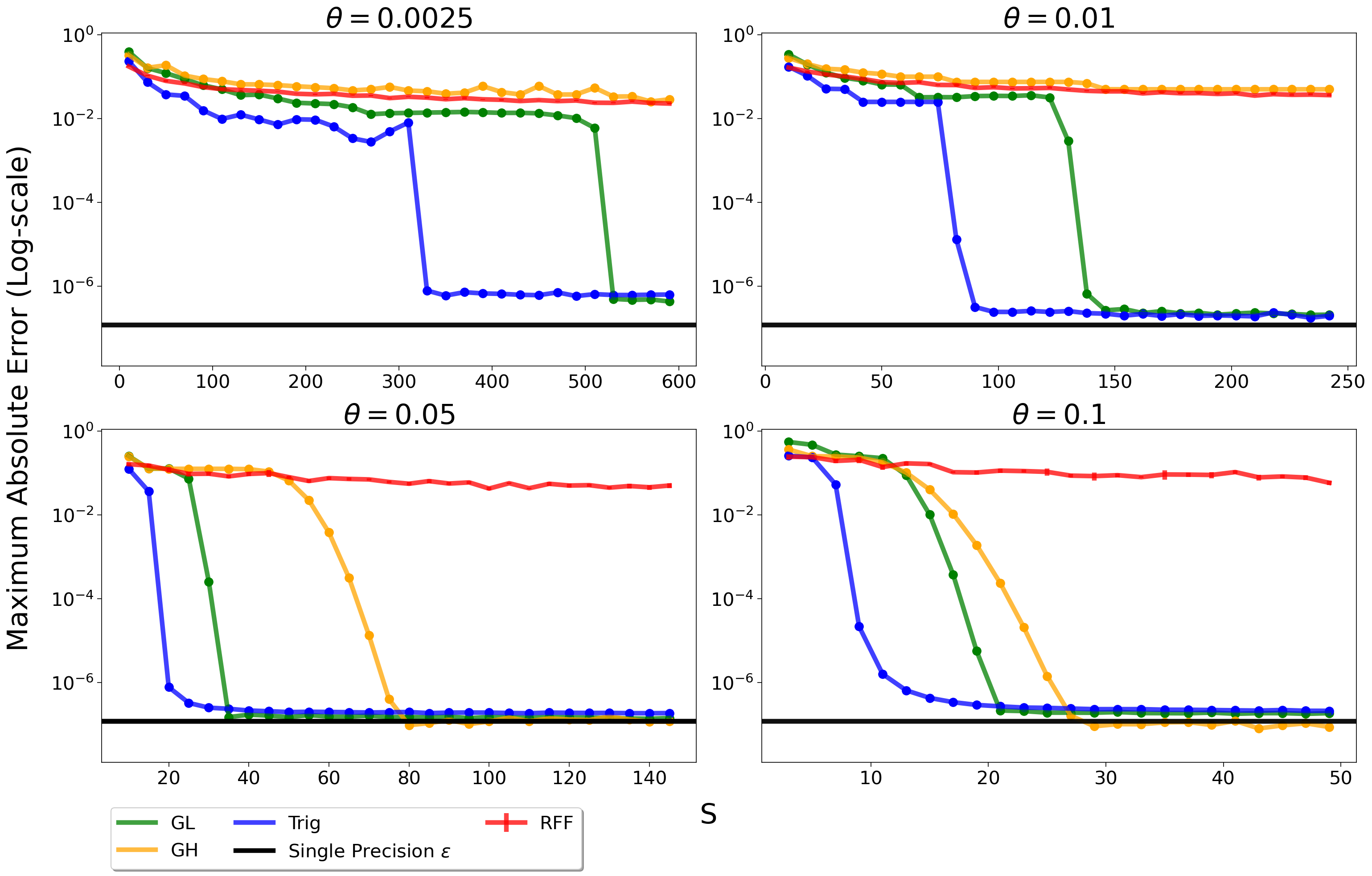}
    \caption{Averaged absolute errors of various Fourier feature maps with $S$ features in approximating the $1$-d SE kernel $k_{\bs{\Theta}}(\tau)$. This average is taken over a grid of $\tau$ values over $[0, 1]$, and $\theta$ is the kernel length-scale.}
    \label{fig:convergence}
\end{figure}

\subsection{Multi-dimensional Extension}
Exploiting the assumption (Assumption \ref{assumptions}(b)) that the spectral density factors across dimensions, we can use tensor product quadrature to extend our $1$-d rule to a trignometrically exact $d$-dimensional rule:

\begin{theorem}[Multi-dimensional TQFF] \label{prop:multi}
Let $\{ (\omega_{i, j}, a_{i, j}) \}_{i=1}^L$ denote the $L$-point trignometrically  exact  quadrature rules in dimension $j$, $j = 1, \dots, d$, as defined by Proposition \ref{prop:tqff}. Define new nodes $\omega_{-i, j} = -\omega_{i, j}$ associated with weights $ \ a_{-i, j} = a_{i, j}$ for all $j = 1, \dots, p$. Let $\mathcal{S}$ be the largest set of multi-indices $\bm{i} = (i_1, \dots, i_d), i_k \in \{-L, \dots, L\}$ such that $\mathcal{S} \subset \prod_{j=1}^d \{-L, \dots, L\}$ but $\bm{i} \in \mathcal{S} \implies -\bm{i} \notin \mathcal{S}$. 
Then, with $a_{\bm{i}} = \prod_{j = 1}^d \frac{1}{2} a_{i_j, j}$ and $\bs{\omega}_k = (\omega_{\bm{i}_1, 1}, \dots \omega_{\bm{i}_d, d})^T$, the quadrature rule $Q_L(f) = \sum_{\bm{i} \in \mathcal{S}} 2 a_{\bm{i}} f( \bs{\omega}_{\bm{i}})$
has trigonometric exactness of degree $2L - 1$. 
\end{theorem}

Maintaining trigonometric exactness of degree $2K-1$ in $d$ dimensions requires ${(2K)^{d}}/{2}$ total points. Because of this, our method suffers the same curse-of-dimensionality present in other tensor product quadrature methods, and should thus be applied only to applications in lower dimensions or with lowwer-dimensional structures, e.g., GPs with additive kernels \citep{duvenaud2011additive, lu2022additive}.

\subsection{Error Bound for Kernel Approximation}
We now provide uniform bounds on  the approximation error from TQFF:
\begin{theorem}[TQFF Error Bound] \label{prop:error}
Let $\bs{\Phi}(\bm{x}), \bm{x} \in \left[0, 1\right]^d$ be the feature map derived from the quadrature rule that has trigonometric exactness of degree $2L - 1$ in each dimension. Define $M = \lceil \frac{\gamma}{\min_{j} \bs{\theta}_j} \rceil $. Let $C_{d}(\bs{\Theta}) = g(\bs{\Theta}) d2^{d-1}$. Then, for any $\bm{x}, \bm{x}' \in \left[0, 1\right]^d$:
\begin{align*}
&|k_{\bs{\Theta}}(\bm{x}, \bm{x}') - \bs{\Phi}(\bm{x})^T \bs{\Phi}(\bm{x}')| \leq 2C_{d}(\bs{\Theta}) \int_{\pi}^{\infty} p_{\gamma}(\gamma \omega) \\
    &+ C_{d}(\bs{\Theta})  \frac{ \left[\pi + 4 + 2\ln\left(\frac{2}{\pi} (4L - 1) \right) \right] \max\{M, 2L- 1\}!}{2 (2L)^{\max\{1, 2L - M\}} (M - 1)!}.
\end{align*}
\end{theorem}
\mycomment{The proof in the supplementary material is very similar to proving the error of standard Gaussian quadrature \citep{Mutny, num_text}, but we use trigonometric analogues for polynomial interpolation \citep{trig_hermite, trig_remainder, trig_remainder2}.} The first term in this bound is the truncation error, and the last term captures quadrature error for the truncated integral. We note that this truncation error rapidly decays with $\gamma$. For example, with the SE kernel and its associated Gaussian spectral density, this truncation error can approximately be bounded by floating point single precision at $\gamma = 1.15$.

Another interesting observation is that the TQFF error bound depends only linearly on the minimum inverse length-scale $1/\min_j\{\bs{\theta}_j\}$, rather than quadratically as in the GHFF bound in \cite{Mutny, dao_quad}. Although the TQFF approximation error decays much faster than RFF, this decay is asymptotically slightly slower than the exponential decrease obtained via Gaussian quadrature. However, we shall see that, empirically, this shortcoming is insignificant in the single precision setting prevalent in machine learning. 

Explicit error bound comparison with GLFF \citep{gl_qff} is difficult due to their unique measure of convergence, and we instead compare these errors empirically. Figure \ref{fig:convergence} shows the average absolute error\footnote{We provide empirical analysis of maximum absolute error plots and error distributions in Appendix \ref{sup:error_dist}.} of these methods when approximating the SE kernel $k_{\bs{\Theta}}(\tau)$ for $\tau \in [0,1]$. Here, $\gamma$ is set at $1.15$ for both TQFF and GLFF so that their error approximately converges to floating point single precision (single precision is used here as it is the default for popular GP implementations \citep{gpytorch, gpflow}, is computationally efficient, and produces similar accuracy to double precision \citep{maddox2022low}). We see that, for each length-scale setting, the average error of the proposed TQFF converges quickest over all methods, with the next best method (GLFF) requiring at least $\approx 50\%$ more features to achieve errors near single precision $\epsilon$. Furthermore, TQFF yields significantly smaller average errors throughout the pre-convergence period. GLFF, on the other hand, yields higher average error than RFF and TQFF for low $S$. RFF can be seen to converge slowly, and GHFF struggles with lower length-scales, as expected. This discrepancy between floating point precision and TQFF/GLFF convergence can be attributed to numerical errors in computing quadrature rules \citep{quad_error}. GHFF does not suffer as heavily from these errors due to the implementation of specialized algorithms for computing Gauss-Hermite rules \citep{townsend2016fast}.  

\section{NUMERICAL EXPERIMENTS}
\label{sec:exp}

We empirically evaluate TQFF against existing Fourier feature approximations for GP regression. Such comparisons are focused primarily on Fourier feature approximation methods, as they possess properties uniquely desirable in a wide range of applications. We do, however, include the stochastic variational inducing points GP approach (SVGP; \citep{hensman_svgp}) as a standard baseline. To compare against the popularly-implemented GHFF approximation, all methods in this section will use an anisoptric SE kernel. All models are trained using the Adam optimizer in PyTorch \citep{kingma2017adam, pytorch}. For GLFF and TQFF, $\gamma$ is fixed at $1.15$ to bound truncation error at single precision. Discussion of the effect of $\gamma$ on kernel approximation error can be found in Appendix \ref{sup:gamma}.

\subsection{2-d Synthetic Functions from GP}
\begin{figure}
    \centering
    \includegraphics[width=\columnwidth]{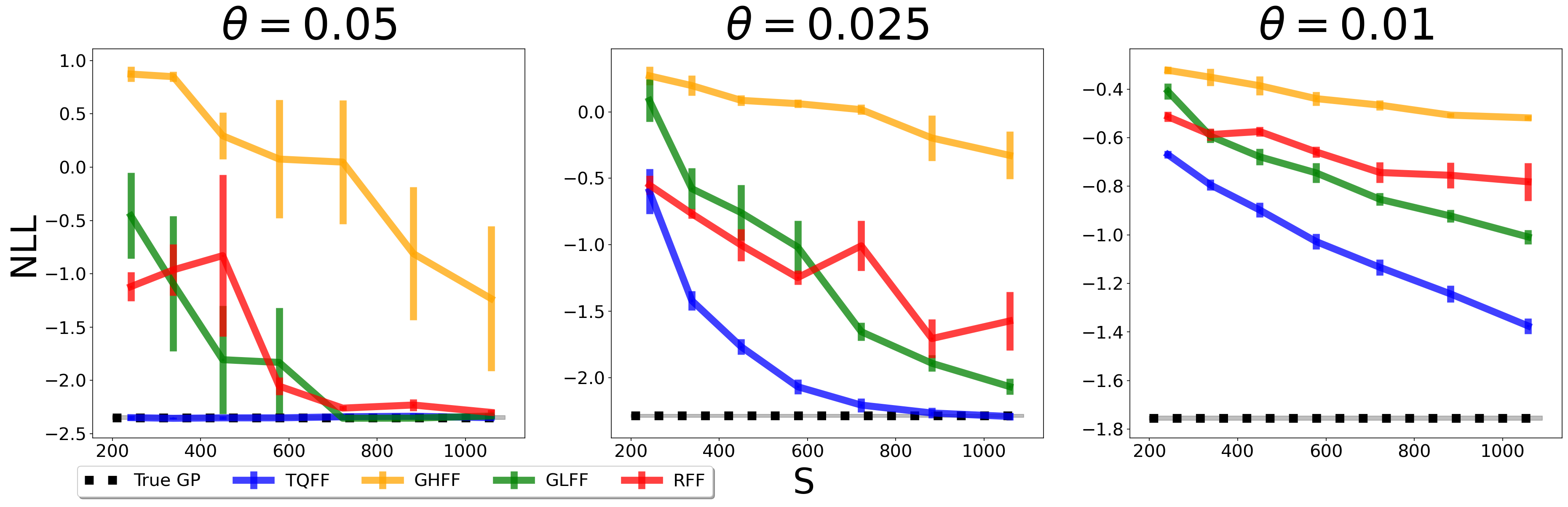}
    \caption{Test NLL for the compared methods for the $2$-d synthetic function sampled from a GP prior. Error bars indicate $\pm$ 1 standard error over 5 random seeds.}
    \label{fig:vary_lengthscales}
\end{figure}

We explore here the effectiveness of TQFF for 2-d synthetic functions. We first generate training data of sample size $n = 20,000$, using functions simulated from a 2-d GP prior with isotropic SE kernels, with different length-scales $\theta = .05, .025$ and $.01$. Predictions are made on test sets of $4,000$ samples. We then compare the learned Fourier feature approximations to a full GP with hyperparameters set as the ground truth. We evaluate the test negative-log-likelihood (NLL) of the compared methods for various feature sample sizes $S$. This procedure is then replicated 5 times.

Figure \ref{fig:vary_lengthscales} shows the test NLL of each method for different length-scales $\theta$. We see that TQFF outperforms competing methods for all $\theta$ and $S$. For $\theta = .05$, TQFF requires noticeably less features than GLFF and RFF to achieve comparable performance to the full GP. GLFF also performs relatively well for large $S$, but is often outperformed by RFF for small $S$. None of the compared methods matches the full GP for $\theta = .01$, but TQFF comes closest by a wide margin. 

\subsection{Approximation of Posterior Uncertainty}

\begin{figure}
    \centering
    \includegraphics[width=\columnwidth]{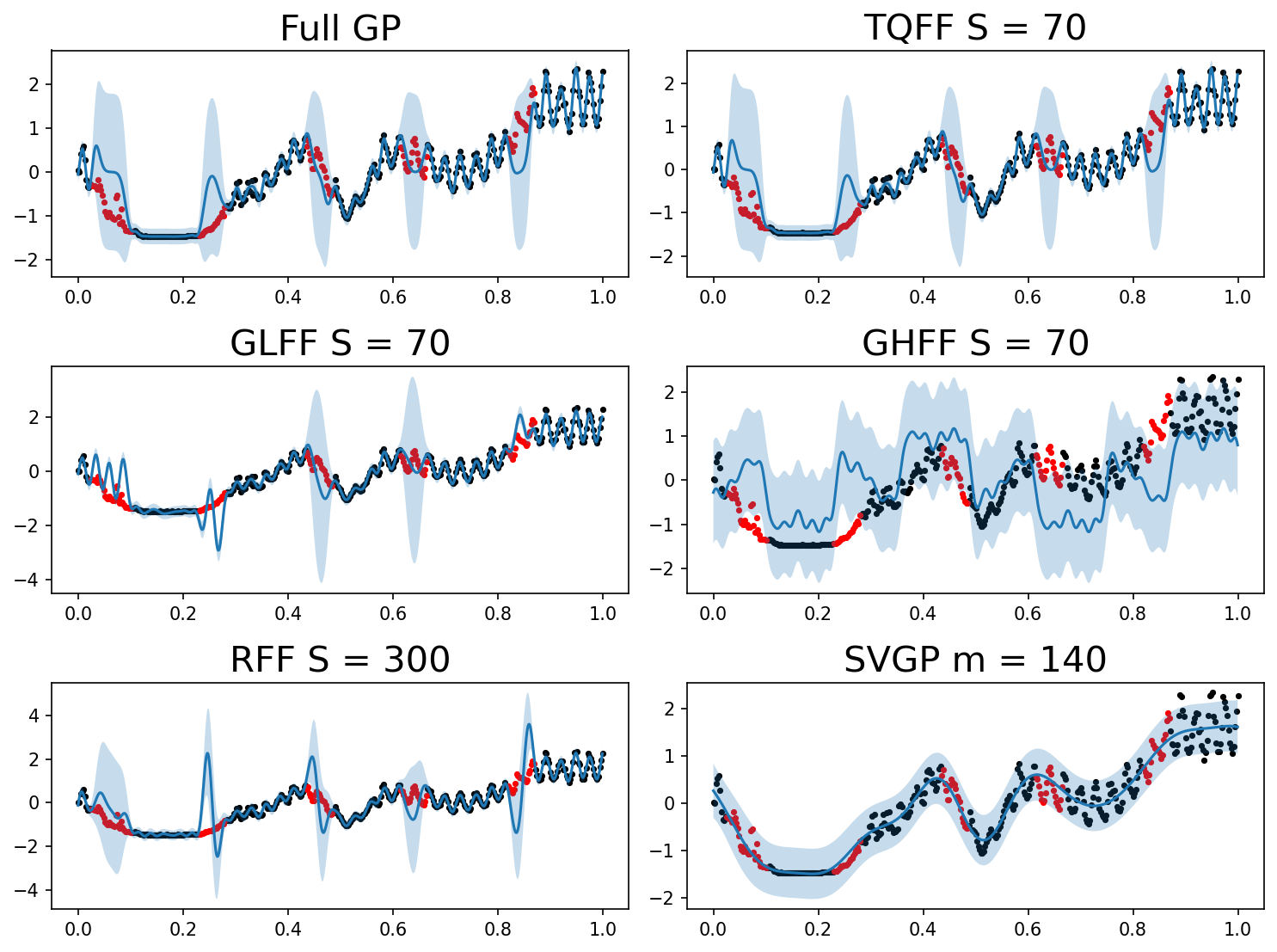}
    \caption{Posterior predictive means and 95\% predictive intervals for the compared methods in the solar irradiance application. The training data is in marked in black, while the hold-out data is in red.}
    \label{fig:solar}
\end{figure}

Next, we examine the performance of TQFF for approximating GP posterior uncertainty, which is crucial for applications such as Bayesian optimization \citep{chen2023hierarchical} and surrogate modeling \citep{li2023additive,ji2022multi}. We adopt the solar irradiance reconstruction experiment in \cite{lean1995reconstruction, gal_2015, vff}, where we removed 5 segments from a time series dataset (representing solar irradiance) and examined the predictive distribution in these hold-out segments (see Figure \ref{fig:solar}). The same methods as before are compared here, with the quadrature Fourier feature approximation methods using $S= 70$ features and RFF using $S = 300$ features. The SVGP baseline is fit with $m = 140$ inducing points.

Figure \ref{fig:solar} shows the mean predictions of the compared methods, with its $95\%$ predictive intervals. We see the posterior predictive distribution from our TQFF is visually indistinguishable from the desired full GP posterior. GLFF and RFF perform well in regions with training data, but suffers from variance starvation in regions with hold-out data, resulting in notable undercoverage. GHFF performs poorly here, due to the aforementioned difficulty of Gauss-Hermite quadrature at low length-scales. SVGP appears to over-smooth the true function in regions of high oscillations, which is undesirable.

\begin{figure}
    \centering
    \includegraphics[width = \columnwidth]{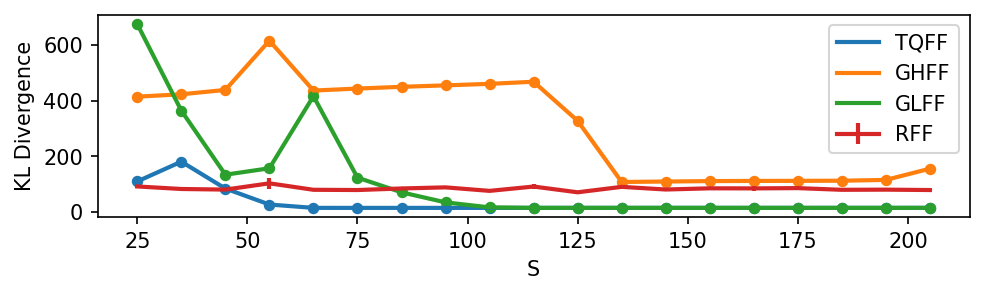}
    \caption{KL-divergences of the predictive distribution of the full GP from the compared approximation methods,at hold-out data points in the solar irradiance application. Error bars for RFF indicate $\pm 1$ standard error over 5 random seeds.}
    \label{fig:solar_kl}
\end{figure}

We can quantify this performance by examining the KL-divergence of the full GP predictive distribution on the hold-out points from the predictive distribution generated from the Fourier Feature approximations. Figure \ref{fig:solar_kl} shows this KL divergence as a function of the number of features $S$, where RFF results are averaged over 5 random seeds to account for sampling variation.  TQFF yields lower KL-divergence from the full GP with far fewer features than the other methods. The KL-divergence of RFF and GHFF appear to converge slowly in $S$, while GLFF requires many features ($S > 100$) to achieve near-zero KL-divergence. 

\begin{figure*}[!h]
    \centering
    \includegraphics[width = \textwidth]{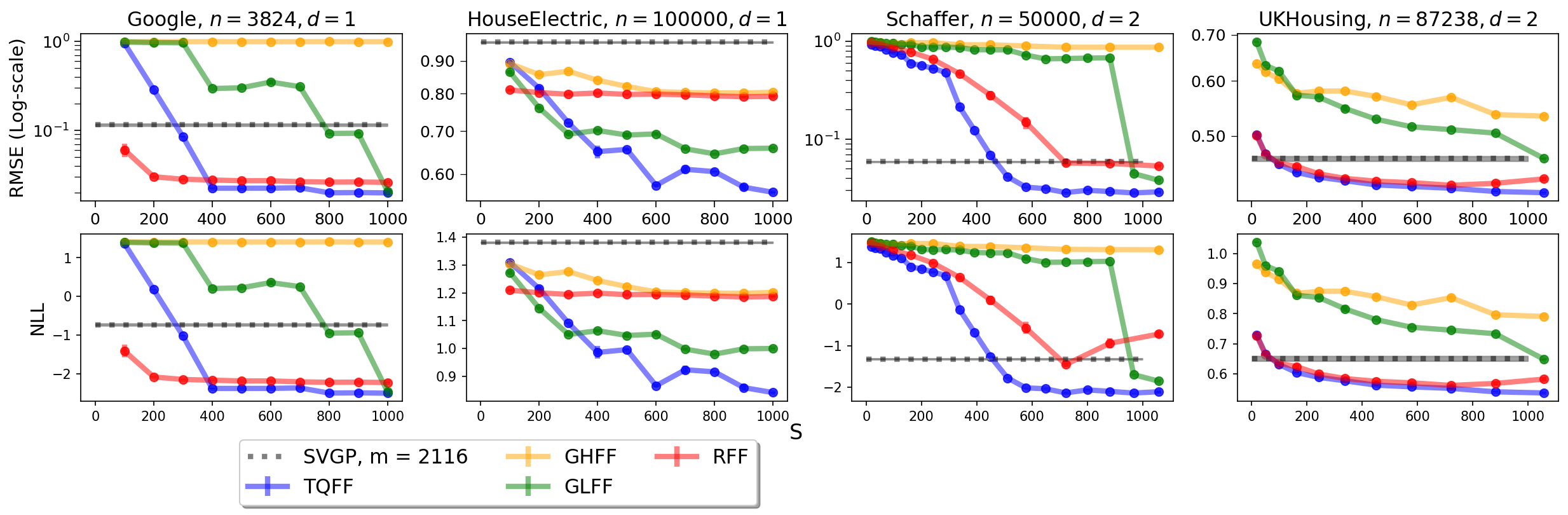}
    \caption{Average test RMSE and NLL on different regression benchmark datasets. Error bars indicate $\pm 1$ standard error over 5 random seeds. Training sample size $n$ and data dimension $d$ are outlined in plot title.}
    \label{fig:benchmarks}
\end{figure*}

\subsection{Regression Benchmarks}

We compare the TQFF on several commonly-used low-dimensional GP regression benchmarks. We examine the time-series dataset of Google daily stock prices \citep{ton, gl_qff}, a household electricity consumption dataset \citep{houselectric}, a UK Apartment Housing price dataset \citep{hensman_svgp}, and the commonly-used Schaffer function benchmark \citep{simulationlib}. Each dataset is randomly split into $80\%$ training and $20\%$ testing. We examine the performance of the Fourier approximation methods up to $S = 1058$ features. The SVGP is again included (using $m = 2116$ inducing points) as a standard benchmark. Details on datasets can be found in Appendix \ref{sup:data}.

Figure \ref{fig:benchmarks} shows the RMSE and NLL over 5 random seeds as a function of the number of features $S$. We see that, for small $S$, RFF may outperform the compared quadrature approximation methods. Across all datasets (and particularly for larger $S$), the proposed TQFF yields the lowest RMSE and NLL among the considered methods and ultimately offers the best overall performance. We also see that, for $d=2$, the feature efficiency of TQFF relative to Gaussian QFF increases for larger $S$, as the number of total nodes scales quadratically with the size of the $1$-d rules. This is consistent with findings in \cite{gl_qff}, who noted that GLFF often requires significantly more than a thousand features before outperforming RFF on real data with low length-scales.

It may seem peculiar that RFF outperforms QFF methods for smaller $S$, when Figure \ref{fig:convergence} shows that QFF often has lower average kernel approximation error. This may be explained by the error distribution discussed in Appendix \ref{sup:error_dist}, which shows that QFF methods can have larger error tails for small $S$. Regardless, the lower average error for the proposed TQFF allows for improved performance over its Gaussian QFF counterparts for smaller $S$, and its fast convergence enables superior performance over RFF for larger $S$.

\section{DISCUSSION}

We proposed a new trigonometric quadrature Fourier feature (TQFF) method for scalable GP modeling. The key idea behind TQFF is  the use of a novel trigonometric quadrature rule, specifically tailored for the desired Fourier transform in Fourier feature approximation. In doing so, this addresses the known limitation of variance starvation for existing Fourier feature methods in GP approximation. We provide approximation error bounds for TQFF, and then demonstrate the improved performance of TQFF over competing methods in a suite of numerical experiments and applications. In particular, we show the TQFF enjoys accurate approximations (with well-calibrated uncertainties) for GPs over a broad range of length-scales using fewer features. 


Our promising results suggest a new class of Fourier feature maps that can be derived from custom interpolants and integrands for the desired Fourier transform integrand. An interesting future direction would be the use of Bayesian quadrature with a kernel designed for the trignometric integrand, to achieve better accuracy in higher dimensions. Applying recent methods for highly-oscillatory quadrature \citep{deano2017computing} may also be fruitful for improving accuracy.

A drawback of the TQFF (along with general QFF approaches) is the curse-of-dimensionality. A potential solution might be to extend the TQFF to higher dimensions via  sparse grid quadrature, as explored in \cite{dao_quad}. This extension would benefit from the improved efficiency of TQFF over Gaussian QFF, and we will explore this as future work. The extension of TQFF for problems with inherent low-dimensional structure is also of great interest, particularly via additive kernels \citep{duvenaud2011additive, lu2022additive}.

\bibliography{bib}

\appendix

\onecolumn
\aistatstitle{Supplementary Material}

\section{Tensor Product Quadrature Rule}
\label{sup:tensor_product}
The 1-dimensional quadrature rules can be extended to higher dimensions via tensor products. If we assume that a multi-dimensional integral factors across dimensions and we apply a $L$ point quadrature rule in each dimension, we can write:
\begin{align*}
&\int_{a}^b p(\bs{\omega}) h(\bs{\omega}) d\omega =  \prod_{j=1}^d \int_{a}^{b} p^{(j)}(\bs{\omega}^{(j)}) f^{(j)}(\bs{\omega}^{(j)}) d\omega^{(j)} \\
&\approx \prod_{j=1}^d \sum_{l=1}^L a_{l,j} h(\omega_{l, j}) = \sum_{\bm{l} \in \prod_{j=1}^d \{1 \dots L\}} a_{\bm{l}} h(\bs{\omega}_{\bm{l}})
\end{align*}
where $a_{l, j}, \ \omega_{l,j}$ are the $l$th quadrature node and weight in dimension $j$ respectively. $a_{\bm{l}} = \prod_{j= 1}^d a_{\bm{l}^{(j)}, j}$ and $\bs{\omega}_{\bm{l}} = (\omega_{\bm{l}^{(1)}, 1}, \dots \omega_{\bm{l}^{(d)}, d})^T$. Clearly, the number of quadrature nodes grows exponentially with dimensions, which limits tensor product quadrature to problems in low dimensions or with low dimensional structure.

\section{Toy Example Details}
\label{sup:toy}

In the example in Section \ref{sec:adv_disadv}, we draw $n = 5000$ samples from the set-up 

\begin{align*}
y_i = \exp(-x_i^2)\exp(\sin(10(x_i - .5))^2) + 3x_i + \epsilon, \epsilon \sim \w{N}(0, .1^2)
\end{align*}

The plots shown in the paper are normalized so that the output has unit standard deviation and zero mean. We draw $x \sim U(0, 1)$ and make predictions for 1000 $x^{*}$ sampled such that $x^{*} \sim U(-1,1)$. GLFF and TQFF are given truncation parameter $\gamma = 1.15$ to bound truncation error at floating point precision.  

\section{Maximum Approximation Error and Error Distribution}
\label{sup:error_dist}
\begin{figure}
    \centering
    \includegraphics[width = \textwidth]{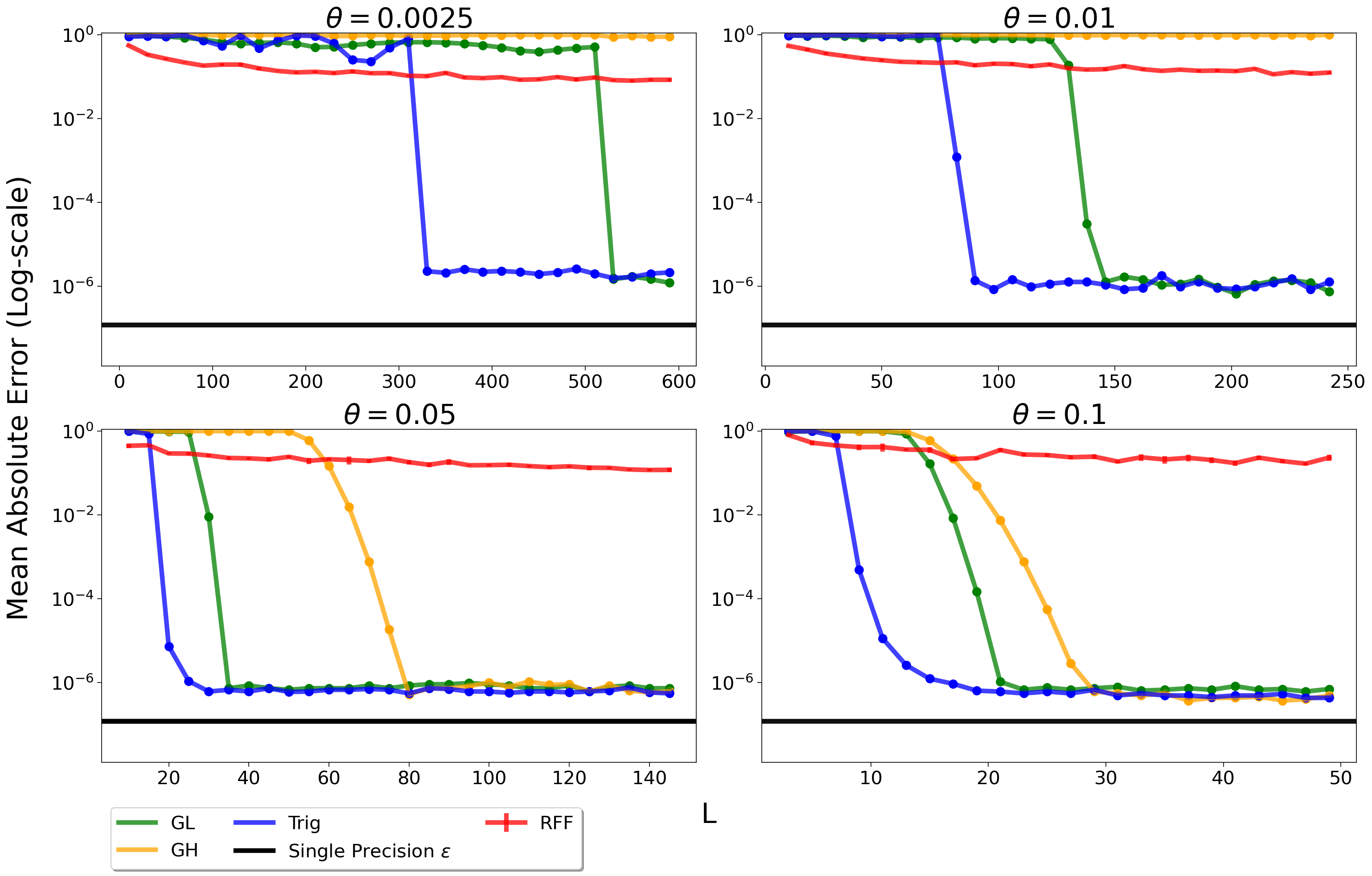}
    \caption{Maximum absolute kernel approximation error for SE $1$d $k_{\bs{\Theta}}(\tau)$.Maximum taken over $n =100$ grid of $\tau$ defined on unit interval.}
    \label{fig:max_error}
\end{figure}

The average kernel approximation error does not tell the full story. Figure \ref{fig:max_error} shows the maximum approximation error of the methods for the SE kernel $k_{\bs{\Theta}}(\tau)$. The maximum is taken over a $n=100$ grid of $\tau$ defined on the unit intereval. For small $L$ we see that $RFF$ performs better. However, only the QFF methods are able to quickly converge to near single precision $\epsilon$ error. The difference between convergence and single precision $\epsilon$ can be attributed to numerical errors for calculating the quadrature rule and kernels. 

\begin{figure}
    \centering
    \includegraphics[width = .5\textwidth]{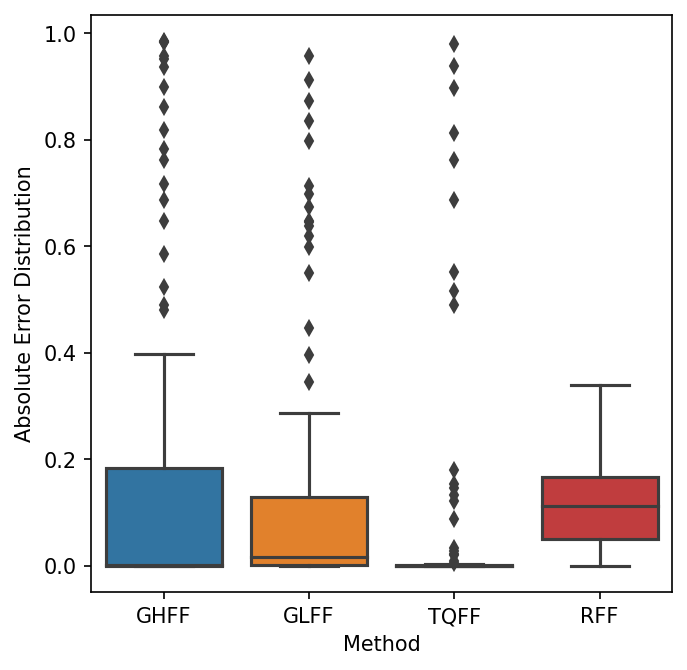}
    \caption{Distribution of absolute errors when approximating an SE kernel $k_{\bs{\Theta}}(\tau)$ for length-scale $\theta = .01$ using $S = 25$ features. $\tau$ is defined over a $n = 100$ grid on the unit interval}
    \label{fig:error_dist}
\end{figure}

We further examine the error distribution for the methods. Figure \ref{fig:error_dist} shows the distribution of absolute errors for the methods when approximating the SE kernel $k_{\bs{\Theta}}(\tau)$ for length-scale $\theta = .01$ using $S = 25$ features. The approximations are made on a grid of $\tau$ over the unit interval. We see that the QFF methods have long tails, while RFF does not suffer from the very large errors. However, we see that the errors of TQFF are highly concentrated around zero, with skinnier error tails relative to the Gaussian QFF methods. TQFF error is also more concentrated around zero than RFF which is consistent with the lower average error we observe.

\section{Effect of $\gamma$ on approximation error}
\label{sup:gamma}
\begin{figure}
    \centering
    \includegraphics[width = \columnwidth]{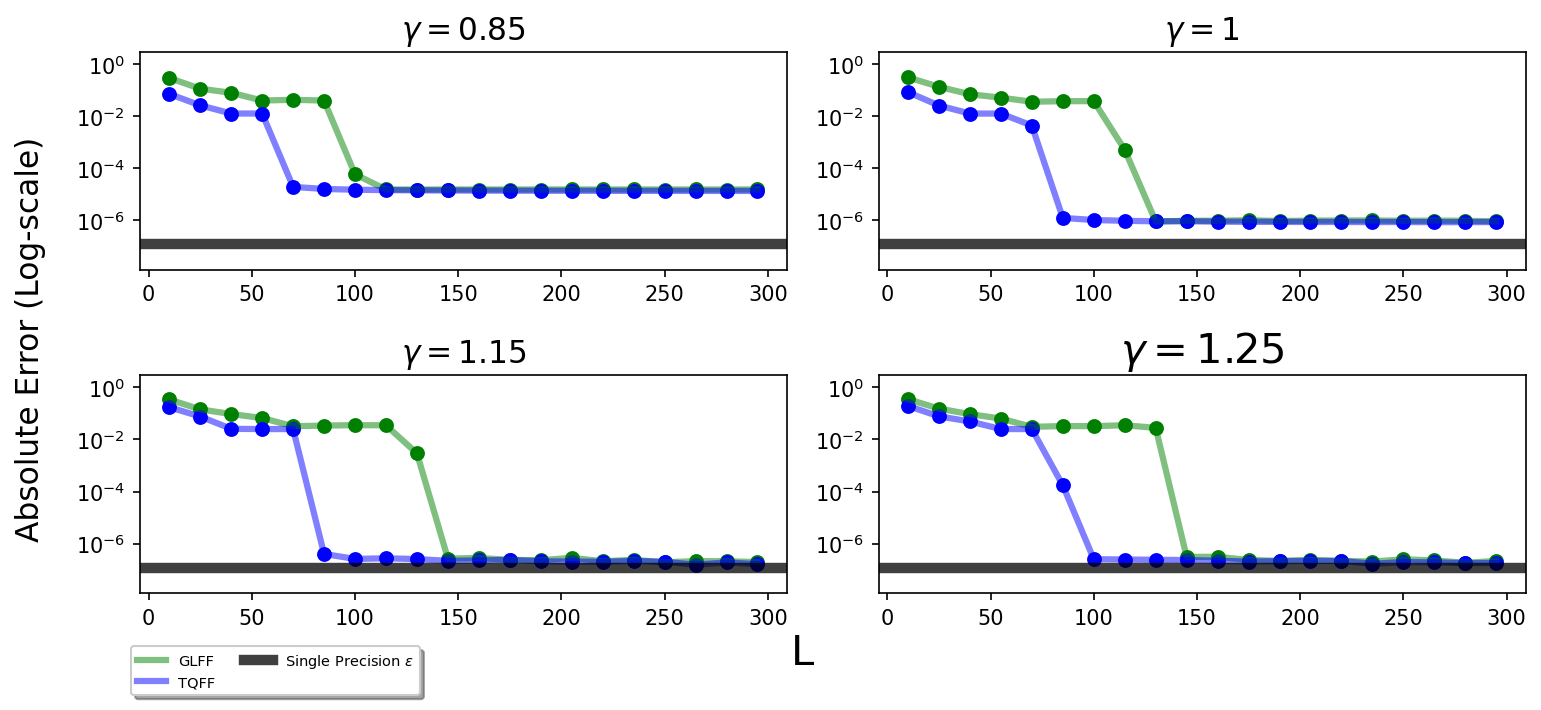}
    \caption{Approximation error as a function of quadrature nodes $L$ for $k_{\bs{\Theta}}(1)$ with $\theta = .01$ for various settings of truncation parameter $\gamma$.}
    \label{fig:gamma}
\end{figure}

We examine the effect of different values of the truncation parameter $\gamma$ on the kernel covariance approximation accuracy of TQFF and GLFF. Figure \ref{fig:gamma} shows the mean absolute approximation error of TQFF and GLFF for the kernel covariance $k_{\bs{\Theta}}(\tau)$. The absolute error is averaged over $\tau$ on a $n=100$ grid on the unit interval given various $\gamma$. TQFF achieves smaller approximation error across values $\gamma$ using significantly fewer features. In addition, as $\gamma$ decreases the approximation error for both methods converges more quickly to the truncation error. This behavior is expected as decreasing $\gamma$ dampens frequency of the oscillatory integrand. However we also see that lower $\gamma$ results in convergence to higher errors due to truncation. The fact that TQFF still performs better across $\gamma$ is expected as dampening the oscillatory behavior benefits all interpolation strategies. 

One could also implement a truncated version of RFF and GHFF to find an optimal balance between quadrature and truncation error for each Fourier Feature approximation. However, this is beyond the scope of this paper and we defer this question to future work. For fair comparison, in the remainder of the paper we set $\gamma = 1.15$ so that the truncation error is upper bounded by approximately single precision machine $\epsilon$. 

\section{Proofs}
\label{sup:proofs}

Note that all the proofs here are in extremely similar flavor to the proofs for standard Gaussian quadrature.A good reference is \cite{num_text}.

\subsection{Uniqueness and Exactness of Interpolating Cosine Polynomial}

Let $P_{L-1}^c(\omega)$ be the degree $L-1$ interpolating cosine polynomial of $f(\omega)$ through points $\{\omega_i\}_{i=1}^L$. Define polynomial function $T_{L-1}(z)$ of degree $L-1$ such that $T_{L-1}(cos(\omega)) = P_{L-1}^c(\omega)$ which is possible as we restrict $\omega \in \left[0, \pi \right)$. Suppose that there exists another polyonmial $H_{L-1}(z)$ such that $T_{L-1}(z_i) = H_{L-1}(z_i)$ for all $z_i = cos(\omega_i), i = 1 \dots n$. Then the polynomial $T_{L-1}(z_i) - H_{L-1}(z_i)$ has $L$ zeros at $cos(\omega_i)$. This is impossible as $T_{L-1}(z) - H_{L-1}(z)$ are degree $L-1$. As $cos(\omega)$ is injective on $\left[0, \pi \right)$ this contradiction shows that there does not exist another degree $L-1$ polynomial function of cosine that interpolates $f(\omega)$ through these $L$ points. 

Uniqueness also implies that $P_{L-1}^c(\omega) = f(\omega)$ when $f(\omega)$ is a cosine polynomial of degree $L-1$.
\subsection{Proof of Proposition \ref{prop:tqff}} 

First we note that for any choice of quadrature nodes $\{\omega_l\}_{l=1}^L$, the an $L$ point quadrature rule will have trignometric degree of exactness $L-1$. To prove exactness note that is $f(\omega)$ is a $L-1$ degree cosine polynomial we can write
\begin{align*}
Q_L(f) = \sum_{i=1}^L a_i f(\omega_i) = \sum_{i=1}^L \int_{-\pi}^{\pi} t^c_i(\omega)f(\omega_i) p_{\gamma}(\gamma \omega) d\omega  = \int_{-\pi}^{\pi} f(\omega) p_{\gamma}(\gamma \omega) d\omega
\end{align*}

As all functions of form $cos(k\omega), k \in \mathbb{N}$ can be written as cosine polynomials the statement follows. Now we  derive the parts of Proposition \ref{prop:tqff}

\subsubsection{Existence of zeros of $q_L^c(\omega)$}

First note that by construction $q_L^c(\omega)$ is orthogonal to all cosine polynomials with degree $< L$, i.e 
\begin{align*}
\int_{-\pi}^{\pi} q_L^c(\omega) p_l^c(\omega) p_{\gamma}(\gamma \omega) d\omega = 0
\end{align*}
For all cosine polynomials $p_l^(\omega)$ with degree $l < L$.  Now we can prove that $q_L^c(\omega)$ has $L$ zeros in $\left[0, \pi \right)$. First by construction we have that
\begin{align*}
    \int_{-\pi}^{\pi} q_L^c(\omega) p_{\gamma}(\gamma \omega) d\omega = 0
\end{align*}

 Therefore $q_L^c(\omega)$ changes sign for at least one $\omega$ in $\left[0, \pi \right)$. Because $q_L^c(\omega)$ is a cosine polynomial, it has at most $L$ zeroes in  $\left[0, \pi \right)$. Assume that $q_L^c(\omega)$ has $1 \leq m < L$ distinct zeroes on $\left[0, \pi \right)$ located at $\omega_1 \dots \omega_m$ of odd multiplicity (aka when $q_L^c(\omega)$ changes sign). Define the degree $M$ cosine polynomial:

\begin{align*}
Z_L(\omega) = \prod_{i= 1}^M (cos(\omega) - cos(\omega_i))
\end{align*}

By construction $Z_L(\omega)q_L^c(\omega)$ does not change sign on the integration interval and therefore $\int_{-\pi}^{\pi} Z_L(\omega)q_L^c(\omega) p(\gamma \omega) \neq 0$. However, this is a contradiction as by assumption $q_L^c(\omega)$ is orthogonal to all cosine polynomials with degree $< L$. Therefore $q_L^c(\omega)$ has at least $L$ zeroes on $\left[0, \pi \right)$. But because it is a cosine polynomial it has at most $L$ zeroes there, so we are done. 

\subsubsection{Exactness}
Let $f(\omega)$ be a cosine polynomial of degree $2L -1$. We can use standard polynomial division to obtain 
\begin{align*}
f(\omega) = q_L^c(\omega) q(\omega) + r(\omega)
\end{align*}
Where $q(\omega), r(\omega)$ are cosine polynomials of degree $\leq L - 1$.  Therefore we write the integral 
\begin{align*}
&\int_{-\pi}^{\pi} f(\omega) p_{\gamma}(\gamma \omega) d\omega  = \int_{-\pi}^{\pi} q_L^c(\omega) q(\omega) p_{\gamma}(\gamma \omega) d\omega + \int_{-\pi}^{\pi} r(\omega) p_{\gamma}(\gamma \omega) d\omega \\
&= \int_{-\pi}^{\pi} r(\omega) p_{\gamma}(\gamma \omega) d\omega 
\end{align*}

Define our $L$-point trignometrically exact quadrature rule with nodes located at the zeroes of $q_L^c(\gamma \omega)$ in $\left[0, \pi \right)$. We have that $Q_L(r(\omega)) = \int_{-\pi}^{\pi} r(\omega) p(\gamma \omega) d\omega$ immediately by the exactness of L-point quadrature. Because $q_L^c(\omega_i) q(\omega_i) = 0$ by the choice of quadrature nodes:
\begin{align*}
&\int_{-\pi}^{\pi} f(\omega) p_{\gamma}(\gamma \omega) d\omega  = \int_{-\pi}^{\pi} r(\omega) p_{\gamma}(\gamma \omega) d\omega  = \sum_{i=1}^L a_i r(\omega_i) = \sum_{i = 1}^L a_i (q_L^c(\omega_i) q(\omega_i) + r(\omega_i)) \\
&= \sum_{i = 1}^L a_i f(\omega_i) = Q_L(f)
\end{align*}
Which shows exactness.

\subsubsection{Positivity of $a_i$}

Recall our definition of $a_i = \int_{-\pi}^{\pi} t^c_{i} p(\gamma \omega) d\omega$. If we employ our quadrature rule of exactness degree $2L - 1$ we have that 
\begin{align*}
\int_{-\pi}^{\pi} (t^c_{i})^2 p_{\gamma}(\gamma \omega) d\omega = Q_L( (t^c_{i})^2) = \sum_{k=1}^{L} a_k (t^c_k(x_k))^2 = a_i
\end{align*}
because $(t^c_{i}(\omega))^2$ has degree $2L - 2$ so that clearly $a_i > 0$.

\subsection{Proof of Proposition \ref{prop:multi}}

Let $Q_{L}(f)$ be defined as in Proposition \ref{prop:multi} in $d$ dimensions. Suppose $f(\bs{\omega}) = cos(\bs{\omega}^T \bm{k})$ for $\bm{k} \in \mathbb{N}^d$ and $||\bm{k}||_{\infty} \leq L- 1$. $\bs{\omega}^{(j)}$ refers to the $j$-th element of $\bs{\omega} \in \mathbb{R}^d$. This is distinct from the $\bs{\omega}_{\bm{i}} \in \mathbb{R}^d$ where the elements of $\bs{\omega}_{\bm{i}}$ are constructed according to the multi-index $\bm{i}$ as stated in the proposition. 

\begin{align*}
    &Q_L(f) = \sum_{\bm{i} \in \mathcal{S}} 2a_{\bm{i}} f(\bs{\omega}_i) = \sum_{\bm{i} \in \prod_{j =1}^d \{-L, \dots L\}} a_{\bm{i}} \exp(i\bs{\omega}_{\bm{i}}^T \bm{k}) \\
    &= \prod_{j = 1}^d \sum_{i = -L, i \neq 0}^L \frac{1}{2} a_{i, j} \exp(i \omega_{i, j} \bm{k}^{(j)}) = \prod_{j=1}^d \int_{-\pi}^{\pi} p_j(\gamma \bs{\omega}^{(j)}) \exp(i \bs{\omega}^{(j)} \bm{k}^{(j)}) d \bs{\omega}^{(j)} \\
    &= \int_{\left[-\pi, \pi \right]^d} p_{\gamma}(\gamma \bs{\omega}) \exp(i\bs{\omega}^T \bm{k}) d \bs{\omega} = \int_{\left[-\pi, \pi \right]^d} p_{\gamma}(\gamma \bs{\omega}) \cos(\bs{\omega}^T \bm{k}) d \bs{\omega}
\end{align*}

Giving us the desired notion of trignometric exactness. The equality in the second line follows from the enforced symmetry of our nodes in each dimension. We write
\begin{align*}
    &\sum_{i = -L, i \neq 0}^L \frac{1}{2} a_{i, j} \exp(i \omega_{i, j} k_j) =\sum_{i = -L, i \neq 0}^L \frac{1}{2} a_{i, j} (\cos( \omega_{i, j} k_j) + i \sin(\omega_{i, j} k_j)) = \sum_{i = -L, i \neq 0}^L \frac{1}{2} a_{i, j} \cos( \omega_{i, j} k_j)  \\
    &= \sum_{i = 1}^L  a_{i, j} \cos( \omega_{i, j} k_j)  = \int_{-\pi}^{\pi} p_{\gamma}(\gamma \bs{\omega}_j) cos(\bs{\omega}_j k_j) d \omega_j = \int_{-\pi}^{\pi} p_{\gamma}(\gamma \bs{\omega}^{(j)}) \exp(i\bs{\omega}^{(j)} \bm{k}^{(j)}) d \bs{\omega}^{(j)}
\end{align*}

\subsection{Proof of Proposition \ref{prop:error}}

\subsubsection{Necessary Results}

We first state a necessary theorem regarding the error of trignometric interpolation from \citep{trig_remainder, trig_remainder2}

\begin{theorem}[Ivanov 1966]
\label{thm:ivanov}
Let $f(\omega)$ be a $r$- times differentiable function. Then  for any trigonometric polynomial $p_K(\omega)$ of degree $K$ that interpolates $f(\omega)$ at $K+1$ dinstinct points in $\left[ -\pi, \pi \right]$ we have that
\begin{align*}
|f(x) - p_K(x)| \leq \left[ \frac{\pi}{2} + 2 + ln\left(\frac{2}{\pi} (2K +1) \right) \right] \times \frac{\sup_{|z(\omega)| = 1} | f_K^{(r)}(z(\omega))|}{(K+1)^r}
\end{align*}
Where $z(\omega) = e^{i\omega}$  and $p_K^{(r)}(z(\omega))$ is the $r$-th (complex valued) derivative of $f_L(\cdot)$ with respect to $z(\omega)$.
\end{theorem}

And a small modification of a relevant theorem for generalizing 1-dimensional quadrature errors to tensor product quadrature from \citep{Mutny}

\begin{theorem}[Mutny 2018]
\label{thm:mutny}
Let $\bs{\omega} \in \mathbb{R}^d$ and $\bm{k} \in \mathbb{R}^d$. Under the assumptions, suppose that the error of a one dimensional quadrature rule approximation to the integral $\int p_j(\bs{\omega}_j \gamma) \cos(\bs{\omega}_j \bm{k}_j) d\bs{\omega}_j$ can be bounded by $\epsilon$. Then the tensor product quadrature error for $\int p(\bs{\omega} \gamma) \cos(\bs{\omega}^T \bm{k}) d\bs{\omega}$ scales as $\epsilon d2^{d-1}$. 
\end{theorem}
We can write the original integral as:
\begin{align*}
\int p(\bs{\omega}\gamma) \cos(\bs{\omega}^T \bm{k}) d\bs{\omega} = \int p(\bs{\omega}\gamma) \exp(i\bs{\omega}^T \bm{k}) d\bs{\omega} = \prod_{j=1}^d \int p_j(\bs{\omega}_j \gamma) \cos(\bs{\omega}_j\bm{k}_j)
\end{align*}
If we can upper bound the error for approximating each integral in the product of the last inequality of $\epsilon$, we can apply lemma 7 from \cite{Mutny} and the error of approximating the original integral is $d2^{d-1}\epsilon$.

\subsubsection{Proof}

First we examine the one dimensional case. We want to bound the error of the $L$ point quadrature of trigonometric degree of exactness $K = 2L-1$. Error bounds on one dimension can then be extended to the multiple dimensions by observing that the one dimensional quadratures involved in the multidimensional extension are exactly equal in value to the standard $L$ point quadrature and therefore we can apply theorem \ref{thm:mutny}.

First we write the form of our feature map/kernel approximation:
\begin{align*}
k_{\theta}(x, x') &=  \int_{-\infty}^{\infty} p_{\bs{\Theta}}(\omega) \cos(\omega (x- x')) d\omega \approx g(\bs{\Theta}) \int_{-\pi}^{\pi} p_{\gamma}(\gamma \omega) \cos\left(\omega \gamma \left[ \frac{x- x'}{\theta} \right]\right) d\omega \\
&\approx g(\bs{\Theta}) Q_L\left(\cos\left(\omega \gamma \left[ \frac{x- x'}{\theta} \right]\right) \right)d\omega) = g(\bs{\Theta}) \sum_{i=1}^L a_i ( \cos(\omega_i \frac{x}{\theta})\cos(\omega_i \frac{x'}{\theta}) +  \sin(\omega_i \frac{x}{\theta})\sin(\omega_i \frac{x'}{\theta})) \\
& = \bs{\Phi}(\bm{x})^T \bs{\Phi}(\bm{x}')
\end{align*}

Where 
\begin{align*}
\bs{\Phi}(x)_i = \begin{cases}
                     \sqrt{a_i \gamma g(\bs{\Theta})} cos(\omega_i \gamma\frac{x}{\theta})  & \mbox{if} \  i \leq L \\
                    \sqrt{a_{i - L} \gamma g(\bs{\Theta})} cos(\omega_{i-L} \gamma\frac{x}{\theta})  &  \mbox{if} \  L < i \leq 2L 
                \end{cases}
\end{align*}

Accuracy of the feature map approximation is clearly exactly the accuracy of the quadrature. Defining $p_{\gamma}(\omega) = \gamma p(\omega)$, we write
\begin{align*}
k_{\theta}(x, x') =  g(\bs{\Theta}) \int_{-\pi}^{\pi} p_{\gamma}(\gamma \omega) \cos\left(\omega \gamma \left[\frac{x - x'}{\theta}\right]\right) d\omega  + 2 g(\bs{\Theta}) \int_{\pi}^{\infty} p_{\gamma}(\gamma \omega) \cos\left(\omega \gamma \left[ \frac{x - x'}{\theta} \right] \right) d\omega
\end{align*}

From now on, define $\alpha = \gamma \left[ \frac{x - x'}{\theta} \right]$. Given our truncation parameter $\gamma$, we will apply quadrature to the first integral. We can use trigonometric Hermite polynomials (\cite{trig_hermite}, Propositions 4.1 and 4.2) to define a degree $K$ trigonometric polynomial of form
\begin{align*}
p_{K}^t(\omega) = c_0 + \sum_{l = 1}^{K} c_l \cos(l\omega) + d_l \sin(l\omega)
\end{align*}
Such that $p_K^t(\omega_i) = cos(\alpha \omega_i) \ , i = 1 \dots L$ where $\{\omega_i\}_{i=1}^L$ are the zeros of the monic orthogonal trignometric polynomial of degree $L$. $p_K^t(\omega)$ is of degree $K = 2L - 1$ and the terms involving $sin(\omega)$ integrate to zero over the symmetric domain and weighting function. Therefore the integral and can be exactly integrated by our quadrature rule:
\begin{align*}
& \int_{-\pi}^{\pi} p_{\gamma}(\gamma \omega) \cos\left( \alpha \omega \right) d\omega  = \int_{-\pi}^{\pi} p_{\gamma}(\gamma \omega) p_{K}^t(\omega) d\omega + \int_{-\pi}^{\pi} p_{\gamma}(\gamma \omega) (\cos(\alpha \omega) - p_{K}^t(\omega)) d\omega \\
&=  Q_L(cos(\alpha \omega))  + \int_{-\pi}^{\pi} p_{\gamma}(\gamma \omega) (\cos(\alpha \omega) - p_{K}^t(\omega))  d\omega
\end{align*}
The total error becomes
\begin{align*}
&|k_{\theta}(x, x') - \bs{\Phi}(x)^T\bs{\Phi}(x')| = |\gamma g(\bs{\Theta}) \left( \int_{-\infty}^{\infty} p(\gamma \omega) cos(\alpha \omega) d\omega - Q_L(cos(\alpha \omega)) \right) | \\
&= g(\bs{\Theta})  | \int_{-\pi}^{\pi} p_{\gamma}(\gamma \omega) (\cos(\alpha \omega) - p_{K}^t(\omega))  d\omega + 2\int_{\pi}^{\infty} p_{\gamma}(\gamma \omega) cos(\alpha \omega) d\omega| \\
& \leq g(\bs{\Theta}) \left( \int_{-\pi}^{\pi} p_{\gamma}(\gamma \omega)  |cos(\alpha \omega) - p_K^{t}(\omega)| d\omega + 2\int_{\pi}^{\infty} p_{\gamma}(\gamma \omega) d\omega \right)
\end{align*}

We need to bound $|cos(\alpha \omega) - p_K^{t}(\omega)|$. Using theorem \ref{thm:ivanov}, letting $ M = \lceil \frac{\gamma}{\theta} \rceil \geq \alpha$  and setting $r = \max \{ 2L - M, 1  \}$ we have that $p_K^t(\omega)$ also satisfies
\begin{align*}
|f(\omega) - p_{2L-1}^c(\omega)|  \leq \left[ \frac{\pi}{2} + 2 + \ln\left(\frac{2}{\pi} (4L - 1) \right) \right] \times \frac{\sup_{|z(\omega)| = 1} | f_L^{(r)}(z(\omega))|}{(2L)^r}
\end{align*}
Define  $z(\omega) = \exp(i\omega)$ and notice that 
\begin{align*}
cos(\alpha \omega) = \frac{1}{2} \left( \exp(i\omega)^{\alpha}  + \exp(-i\omega)^{\alpha} \right) = \frac{1}{2}(z(\omega)^{\alpha} + z(\omega)^{-\alpha}) := f(z(\omega))
\end{align*}
Note that because we restrict the supremum to $\omega$ such that $|z(\omega)| = 1$, repeated differentiation of $z(\omega)^{\alpha}, z(\omega)^{-\alpha}$ wrt $z(\omega)$ gives us
\begin{align*}
\sup_{|z(\omega)| = 1} | f_L^{(r)}(z(\omega))| \leq  \frac{(\lceil \alpha \rceil + r - 1)!}{ ( \lceil\alpha \rceil - 1)!} \leq \frac{( M + r - 1)!}{ ( M - 1)!}
\end{align*}. 

Assuming that $x, x' \in \left[0, 1\right]$,  $\alpha \leq \lceil \frac{\gamma}{\theta} \rceil = M$. Therefore our final bound can be written, definint $H = \max\{2L - 1, M \}$
\begin{align*}
|k_{\theta}(x, x') - \bs{\Phi}(x)^T\bs{\Phi}(x')| \leq
    g(\bs{\Theta}) \left( \left[ \frac{\pi}{2} + 2 + \ln\left(\frac{2}{\pi} (4L - 1) \right) \right] \times \frac{ \max\{M, 2L- 1\}!}{(2L)^{\max\{1, 2L - M\}} (M - 1)!} + 2\int_{\pi}^{\infty} p_{\gamma}(\gamma \omega) \right) 
\end{align*}

The extension to the multi-dimensional case involves the tensor product of one-dimensional rules that produce identical error to the 1-dimensional quadrature produced here. Therefore we can apply theorem \ref{thm:mutny} and the result follows.

\section{Computation of Nodes and Weights in the Golub Welsh Algorithm}
\label{sup:nodecomp}
Recall that we need to compute a monic cosine polynomial of degree $L$ $q_L^c(\omega)$ that is orthogonal to all cosine polynomials of degree less than $L$. We can use Three term recurrence relation and the Golub-Welsh algorithm for this task \citep{num_text}. The monic orthonormal cosine polynomials $\{q_l(\omega)\}$ associated with the weight function $p(\omega)$ on $\left[a, b \right]$ satisfy the relation:

\begin{align*}
q_1^c(\omega) = (cos(\omega) - B_0)
q_{k+1}^c(\omega) = (cos(\omega) - B_k) - A_k (p_{k-1}(\omega))
\end{align*}

where $A_k = \frac{||q_k^c||^2}{||q_{k-1}^c||^2} ,  \ k \geq 1$ and $B_k = \frac{\langle cos(\omega) q_k,q_k \rangle}{||q_k^c||^2}, k \geq 0$. Where the inner product is $\langle f, g \rangle = \int_{a}^b f(\omega)g(\omega) p(\omega) d\omega$. All inner products involve integrals of powers of $cos(\omega)$ against the weight function $p(\omega)$ and can be calculated analytically using a software such as Mathematica or Maple. Analytical solutions exist for kernels such as the RBF and matern. 

Because cosine polynomials can be written as polynomials of functions defined on the interval $[-1, 1]$,  we can apply the standard Golub-Welsh algorithm using the eigenvalues/eigen vectors of the tri-diagonal matrix formed from $B_k, A_k$ to obtain nodes/weights that satisfy the conditions of Proposition \ref{prop:tqff}. Please see \cite{num_text} or any numerical analysis text for more details. The only difference is we have to take the inverse cosine transformation of the eigenvalues of the matrix to get our nodes. 

Implementation of  Golub-Welsh and TTR requires $\w{O}(L^3)$ complexity to compute $L$ nodes. This computational cost is not too burdensome as the quadrature rules only have to be computed once independent of any dataset. We note that this approach works well for obtaining quadrature rules and fourier features up to $L \approx 1000$ which is the upper bound for most applications. Numerical error begins to accumulate at this point and computation is burdensome. To efficiently produce more features one can easily apply asymptotic type methods from standard Gaussian quadrature to scale to hundreds of thousands of features \citep{townsend2016fast}.

\section{Benchmark Data Sources}

For the google data we obtained the log daily high stock price from \url{https://finance.yahoo.com/quote/GOOG?p=GOOG}. We took data from 9/11/2004 - 9/13/2023. The Household electricity data-set was taken from the frist 125,000 observations from the dataset stored in \url{https://archive.ics.uci.edu/dataset/235/individual+household+electric+power+consumption}. UKHousing data was obtained from the 2018 price paid dataset of sale prices filtered for flats (apartments) located at \url{https://www.gov.uk/government/statistical-data-sets/price-paid-data-downloads}. We extracted the prices from May-December. The Schaffer function is a widely used benchmark function. Specification can be found at \url{https://www.sfu.ca/~ssurjano/schaffer2.html}. We evaluate the function on the hyercube $[-3, 3]^2$. 

\label{sup:data}

\end{document}